\documentclass[10pt, a4paper]{article}

\usepackage{lrec-coling2024} 
\usepackage{listings}
\usepackage{amsmath}

\usepackage{natbib}
\usepackage{multibib}
\makeatletter
\def\@mb@citenamelist{cite,citep,citet,citealp,citealt,citepalias,citetalias}
\makeatother
\newcites{languageresource}{~}

\usepackage{graphicx}
\usepackage{tabularx}
\usepackage{soul}
\usepackage{array}
\usepackage{titlesec}

\usepackage{xcolor}
\definecolor{mahogany}{rgb}{0.75, 0.25, 0.0}

\titleformat{\section}{\normalfont\large\bfseries\center}{\thesection.}{1em}{}
\titleformat{\subsection}{\normalfont\SmallTitleFont\bfseries\raggedright}{\thesubsection.}{1em}{}
\titleformat{\subsubsection}{\normalfont\normalsize\bfseries\raggedright}{\thesubsubsection.}{1em}{}
\renewcommand\thesection{\arabic{section}}
\renewcommand\thesubsection{\thesection.\arabic{subsection}}
\renewcommand\thesubsubsection{\thesubsection.\arabic{subsubsection}}

\usepackage{xcolor}
\usepackage{hyperref}
 \definecolor{darkblue}{rgb}{0, 0, 0.5}
  \hypersetup{colorlinks=true, citecolor=darkblue, linkcolor=darkblue, urlcolor=darkblue}

\usepackage{xstring}

\usepackage{color}

\title{FRENCH-YMCA: A FRENCH Corpus meeting the language needs of Youth, froM Children to Adolescents.}

\name{
\parbox{\textwidth}{\centering
Cherifa Ben Khelil$^{1,2}$, Jean-Yves Antoine$^{2}$, Anaïs Halftermeyer$^{3}$, Frédéric Rayar$^{2}$,\\
and Mathieu Thebaud$^{4}$ \\[3mm]
}
}

\address{
\parbox{\textwidth}{\centering
$^{1}$EFREI Research Lab - University of Paris Panthéon Assas, France\\
$^{2}$LIFAT - University of Tours, France\\
$^{3}$LIFO - University of Orleans, France\\
$^{4}$CMRRF, Mutuality of Morbihan, France\\[1mm]
Emails: cherifa.ben-khelil@efrei.fr, Jean-Yves.Antoine@univ-tours.fr,\\
anais.halftermeyer@univ-orleans.fr, frederic.rayar@univ-tours.fr,\\
mathieu.thebaud@vyv3.fr
}
}
\abstract{In this paper, we introduce the French-YMCA corpus, a new linguistic resource specifically tailored for children and adolescents. The motivation for building this corpus is clear: children have unique language requirements, as their language skills  are in constant evolution and differ from those of adults.  With an extensive collection of 39,200 text files, the French-YMCA corpus encompasses a total of 22,471,898 words. It distinguishes itself through its diverse sources, consistent grammar and spelling, and the commitment to providing open online accessibility for all. Such corpus can serve as the foundation for training language models that understand and anticipate youth's language, thereby enhancing the quality of digital interactions and ensuring that responses and suggestions are age-appropriate and adapted to the comprehension level of users of this age.
\newline  
}

\begin{document}

\maketitleabstract
\section{Introduction}\label{sec1}
The advent of Artificial Intelligence (AI) has fundamentally transformed how we engage with Natural Language Processing (NLP). Language models driven by neural networks have elevated AI capabilities to high levels, enabling a diverse range of applications, from machine translation to text generation.

With the emergence of machine learning techniques, the issue of adaptation is now part of everyday considerations in NLP. In particular, pre-trained large language models are usually fine-tuned through transfer learning on specific data, following supervised, semi-supervised or unsupervised approaches. 
Model adaptation has, so far, been approached from various angles, including specialized language (such as medical terminology), modality (written vs. spoken language), language register, and even the diachronic evolution of language \cite{blouin:hal-03550384}.
Amid the sources of variation requiring an adaptation of these models, the speakers' language skills has been largely neglected. For instance, to the best of our knowledge, little effort has been directed towards adapting models for individuals with dyslexic disorders, possibly due to the insufficient availability of relevant target data.
Likewise, there is a growing need for linguistic resources that suit the specific characteristics of children's language. Existing language models are frequently trained on general training data that overlook the 
language development from child ages. Hence, text prediction, chatbots, or voice assistance tools that present a satisfactory behaviour for adults often face significant shortcomings when dealing with younger users. 
In this paper, we address the need for a French corpus specially designed for young users. In Section \ref{sec2}, we examine existing French speaking child corpora and address their limitations. In Section \ref{sec3}, we introduce our response to this challenge: the "French-YMCA" corpus, which stands for French corpus meeting the language needs of \textbf{Y}outh, fro\textbf{M} \textbf{C}hildren to \textbf{A}dolescents. We provide insights into various aspects, including data collection and preprocessing, followed by a statistical analysis of our corpus. In the final section, we highlight the corpus' potential and its applicability, especially in the context of our research domain, which focuses on Augmentative and Alternative Communication (AAC) methods for children with severe sensorimotor disorders.
\section{Related Works}\label{sec2}
Four initiatives can be identified, that constructed French corpora designed for children (Table \ref{tab1}).
One of the notable corpora in this category is CHILDES\footnote{\url{https://childes.talkbank.org/}} \cite{childes}. It offers a collection of corpora in various languages, including French. 
The corpora encompass diverse media formats, such as audio and video recordings. Within the CHILDES French corpora, there is a diverse range of age groups, from infants to early childhood, making them a valuable source for investigating language development, 
However, it should be noted that these recordings fell short in creating a 
relevant resource for training language models tailored for children, due to the fact that the dialogues mainly feature very young children with an extremely limited vocabulary. 

Existing alternative initiatives, E-Calm \cite{ecalm} and Philosophèmes \cite{Philosophemes}, focus on textual data. E-Calm\footnote{\url{https://e-calm.huma-num.fr/}} 
provides access to nearly 4,500 texts, including their preliminary drafts, all typed and annotated. Its objective is to create a comprehensive corpus of educational and academic texts specifically designed for young learners. Nevertheless, it is important to emphasize that 
these writings often contain a multitude of errors, reflecting the ongoing learning process of these young students.

The Philosophèmes\footnote{\url{https://philosophemes.msh.uca.fr/}} database is the result of fifteen years of research applied to education. These datasets were exclusively gathered in regular school. 
This resource provides various types of data, including spoken, multimodal, and written texts organized into four distinct corpora: The "\textit{Philosophèmes}" corpus, the "\textit{Grenouille}" corpus \cite{Grenouille}, the "\textit{C'est pas moi}" corpus \cite{cestpasmoi}, and the "\textit{PréCPhi" (\textbf{Pr}emier \textbf{éc}rits \textbf{Phi}losophiques)} corpus. 
Similar to E-Calm, Philosophèmes share a distinctive feature: its content consists exclusively of essays and productions created by young students. This also means that the texts in the Philosophèmes' corpora contain grammatical and spelling errors. Besides, these texts are limited to the topics chosen by educators due to their educational relevance, resulting in a restricted range of terms and vocabulary usage. 

The last existing resource appears to be more varied and is provided by the TextToKids project \footnote{\url{http://texttokids.irisa.fr/}}. This project is dedicated to simplify the creation and filtering of texts tailored for children, especially when it comes to explaining current events in a way that matches their language skills. As a part of this initiative, a diverse corpus of French texts, primarily oriented toward children between the ages of 6 and 14, was constructed and enriched through an emotion annotation system \cite{etienne-etal-2022-psycho}. This corpus consists of 1,594 texts covering a range of genres including journalism, fiction, and encyclopedic content. However, despite its diversity, its relatively modest size still poses challenges for training large language models. Furthermore, this corpus remains unavailable online, which restricts access for researchers.

In order to overcome the various limitations observed on current existing resources, we propose a new French corpus that answer the following objective :
\begin{itemize}
    \item To provide a large, rich and reliable database that can be used to train language models specifically tailored for youth.
    \item To create a diverse corpus of texts specially designed for youth, including a variety of language registers 
    \item To create a resource with reliable grammar and spelling.
    \item To release the resource under an open license. 
\end{itemize}
\begin{table*}[]
    \begin{center}
    \setlength{\tabcolsep}{2pt}
    \scriptsize
\begin{tabular}{llccc}
\hline
\multicolumn{1}{c}{\textbf{Corpus name}}                     & \multicolumn{1}{c}{\textbf{Text source}}                                                                                                                                                                & \multicolumn{1}{c}{\textbf{Text genre}}                                            & \multicolumn{1}{c}{\textbf{Number of files}} & \multicolumn{1}{c}{\textbf{Age range}} \\ \hline
CHILDES                                                      & {\parbox{5cm}{Transcriptions from audio  and video recordings of children's conversations }}                                                                              & Conversation                                                                       & 2,275                                        & 6 months to 11 years                  \vspace{0.2cm} \\
\hline
E-calm                                                       & Student Essays and writings                                                                                                                                                                             & Formal writing                                                                     & 4,500                                        & 6 to 18 years                         \vspace{0.2cm} \\
\hline
Philosophèmes                                                & {\parbox{5cm}{Transcriptions of philosophical discussion}}                                                                                                                  &Philosophical conversation             & 22                                           & 5 to 18 years                         \vspace{0.2cm} \\
Grenouille                                                   & Student Essays                                                                                                                                                                                          & Formal writing                                                                     & 400                                          & 8 to 14 years                         \vspace{0.2cm} \\
C'est pas moi                                                & {\parbox{5cm}{Transcriptions of discussions on "Lying" and writings }}                                                                                                        & Conversation and  formal writing         & 53                                           & 6 to 8 years                           \\
PréCPhi                                                      & Philosophical Writings                                                                                                                                                                                  & Formal writing                                                                     & 1,000                                        & 8 to 15 years                          \vspace{0.2cm}\\
\hline
\begin{tabular}[c]{@{}l@{}}TextToKids\\  corpus\end{tabular} &  {\parbox{5cm}{Newspapers Albert, and P’tit Libé, articles from different  newspapers, novels for children, and encyclopedic texts, taken from  Vikidia}} & \begin{tabular}[c]{@{}l@{}}Newspaper, \\ encyclopedia, \\ and fiction\end{tabular} & 1,594                                       & 6 to 14 years       \vspace{0.2cm} \\  
\hline
\end{tabular}
\end{center}
\caption{\label{tab1}Overview of varied corpora for children and adolescent.}
\end{table*}
\section{French-speaking corpus for youth languages}\label{sec3}
\subsection{Data collection and preprocessing} \label{sec31}
One of the challenges in building a new French corpus tailored for children and adolescents is to ensure text diversity while maintaining high linguistic quality, taking into account different age groups.

In order to meet this objective, our first step was to collect a large number of relevant texts for children (aged 6-11 years) and adolescents (aged 12-17 years) from various sources, such as e-journals, Vikidia, books, copyright-free stories available online and subtitles for animation movies and cartoons. The content, style and language registers of the corpus were diversified to adequately represent the language patterns and vocabulary utilized by children and adolescents. In particular, we design our corpus to be able to distinguish three well-identified language registers : news, literacy and informal communication. After collecting the texts, we pre-processed them using NLP techniques in order to obtain a cleaner learning corpus. This involved converting letter cases, grouping certain tokens together into generic categories (such as numbers or dates), and removing uncommon words that could potentially lead to errors. 

\subsection{Corpus description}\label{sec32}
\begin{figure}[ht!]
\centering
\includegraphics[scale=0.4]{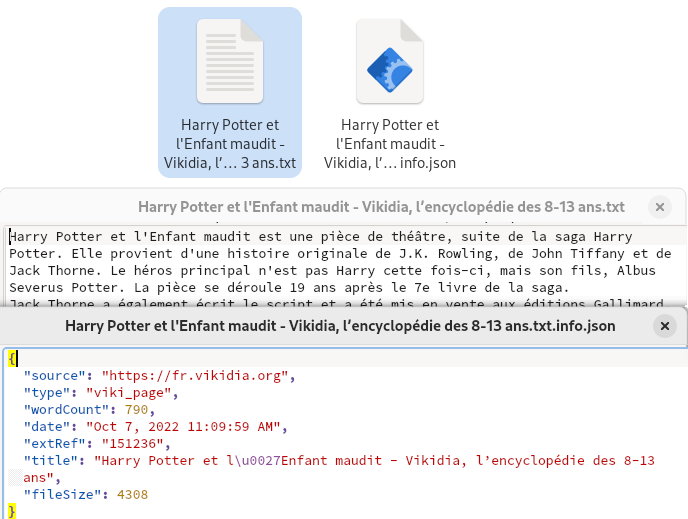}
\caption{Example of a text file and its corresponding JSON metadata file from the French-YMCA Corpus.}
\label{ymca}
\end{figure}

Table \ref{tab2} presents the details of the text sources that were collected and analyzed, providing insight into their characteristics. The corpus is referred to as the "\textit{\textbf{French-YMCA} corpus}". It consists of a set of TXT files, each containing raw text, accompanied by an associated JSON metadata file. This metadata file includes information such as the file name, its source, type, word count, creation date, and size. Figure \ref{ymca} provides a visual of the corpus structure, with a text file and its corresponding JSON metadata.
With a collection of 39,200 text files, the corpus contains a total of 22,471,898 words and an average sentence length of 12 words. The number of files and total word count vary among the different sources in the corpus.

Children's books and stories, the first source type, consists of 2,211 files and a total of 2,935,108 words. These texts address the literary language register. Children's e-journals, the second source type, includes 16,026 files, containing a total of 5,103,070 words. These texts are also intended for children, but concern a specific formal language register relative to newspapers.
Adolescent e-journals, the third source type, includes 2,079 files totaling 949,753 words and using a formal newspaper language register. The fourth source type is Vikidia, a youth encyclopedia with 17,237 files and 7,380,229 words, written in a formal informative language register. The text in Vikidia is appropriate for a broad age range, from elementary school to early high school, making it suitable for educating young reader on a variety of topics. Finally, subtitles for animation movies and cartoons consist of 1,647 files containing 6,103,738 words, intended for children and corresponding to a rather informal language register.
\begin{table*}[ht!]
\centering
\renewcommand{\arraystretch}{1} 
\setlength{\tabcolsep}{1pt} 
    \scriptsize
\begin{tabular}{p{2.75cm} c p{2.5cm} p{1.5cm} p{2cm} p{1.5cm} p{2cm}}
\hline
\textbf{Text source} & \textbf{Age} & \textbf{Language register} & \textbf{Number of files} & \textbf{Number of words} & \textbf{Avg. sentence length} & \textbf{Data availability} \\
\hline
Books and stories for children & Children & Literature & 2 211 & 2 935 108 & 12 & >80\% open; rest by agreement \\
Children e-journals & Children & Formal newspaper & 16 026 & 5 103 070 & 17 & By agreement only \\
Adolescents e-journals & Adolescents & Formal newspaper & 2 079 & 949 753 & 16 & By agreement only \\
Vikidia & Adolescents & Formal informative & 17 237 & 7 380 229 & 20 & Public \\
Animated movie subtitles & Children & Informal & 1 647 & 6 103 738 & 7 & By agreement only \\
\hline
\textbf{Total} & & & 39 200 & 22 471 898 & 12 & \\
\hline
\end{tabular}
\caption{\label{tab2}Distribution of the collected texts based on their source type.}
\end{table*}

The variation in sentence length among the different sources is indicative of the diversity in their writing style and structure. Subtitles for animation movies and cartoons have the shortest average sentence length among the text sources in the corpus. This can be attributed to their purpose of conveying information quickly and efficiently. As such, this sub-corpus can be compared to an informal language register close to spontaneous conversational speech. 
The literature sub-corpus one, on the other hand, has an average sentence length of 12 words, ranking it second in terms of the shortest average sentence length among text sources, following the informal category. This suggests that the writing style of its texts is simple and suitable for young readers. In contrast, texts from the newspaper language register exhibit a more advanced writing style, featuring sentences of greater average length.
The informative sub-corpus had an average sentence length of 20 words, which is longer than that of the other sources, but still within the expected range for texts in an encyclopedia.

Our aim was to create an open-access corpus available to the widest possible audience. However, access to the corpus is regulated under copyright and distribution laws, allowing only texts that are in the public domain or explicitly authorized by the rights holders. About 65\% of the files, representing 75\% of the word count, are publicly accessible under a CC-BY-SA license. The access and use of texts without authorization remains restricted solely for research purposes, under a formal collaboration agreement with the corpus-producing laboratory.

The French-YMCA corpus offers several notable advantages when compared to existing resources (ref. section \ref{sec2}). Firstly, it shares similarities with the corpus of the TextToKids project, as both contain texts from various sources and genres. However, our corpus surpasses it by providing an even more varied range of linguistic resources. This guarantees a wide coverage of topics and writing styles, meeting the diverse interests of young users effectively. Furthermore, our corpus is distinguished by its lack of spelling errors, in contrast to E-Calm and Philosophèmes, which enhances its value for training language models, creating a robust basis for machine learning.  
\subsection{Unlocking the potential of youth-adapted text corpora}\label{sec33}
As we explore the many applications of our corpus, it is essential to highlight its central role as the foundation for training linguistic models. These latter are designed  to understand and predict young people's language, with a primary focus on improving the quality of digital interactions. This technology ensures that the responses and suggestions generated are not only adapted to the user's age but also finely tuned to their comprehension level.
Furthermore, we emphasize the crucial role of prioritizing digital inclusivity, which holds great importance. Children and adolescents should have access to a positive and educational digital experience. Models developed from specialized corpora for youth  contribute significantly to the creation of online environments customized to their needs, ensuring safe navigation, learning, and interaction with technologies that align with their cognitive and linguistic development.
Additionally, employing these language models enhances technology interactions by making them more effortless and user-friendly, which fosters the development of independence and self-confidence among young users when engaging with digital devices.

For example, chatbots, and voice assistant systems created for children are becoming increasingly popular for interacting with young users. However, these systems have faced challenges such as failing to detect signs of potential issues like sexual abuse, eating disorders, and illegal drug use. The importance of employing a youth-specific corpus becomes apparent in such instances, as these technologies must grasp the nuances of young users' queries and provide appropriate, easily comprehensible responses. To achieve this, the AI driving these systems requires training with youth-centric linguistic resources, such our French-YMCA corpus, to ensure their effectiveness.


When focusing on the improvement of accessibility for children and adolescents with disabilities, these dedicated corpora are indispensable in developing digital accessibility tools customized for this demographic, including AAC systems. Precisely, we used our French-YMCA corpus to train next-word prediction models for such systems. A general model based on an adult corpus has already been developed. However, this model sometimes suggests complex or even technical words that are not suitable for youngs users’ reading and comprehension skills. For instance, if a child tries to compose a message in French about wanting to drink something "j’aimerai boire une"/ "\textit{I would like to drink one}", the model based on an adult corpus suggests the word "bière"/ "\textit{beer}" first with a probability of 96.41\%. This suggestion is unlikely to help the child and could even be frustrating. In contrast, when we use the model based on the French-YMCA corpus, it suggests words like "tasse" / "\textit{cup}" with a score of 37.54\% and "gorgée" / "\textit{sip}" with a score of 37.53\%, which align better with a child's language. We carried out a theoretical evaluation of the newly adapted models (anonymous reference) and our findings clearly demonstrated that our models work better than the general one trained on data for adults.

\section{Conclusion} \label{sec4}
In this paper, we introduced the French-YMCA corpus a valuable linguistic resource designed specifically for children and adolescents, filling an important gap in the field of NLP and AI. With its varied content and reliable grammar and spelling, this corpus  holds significant potential in the training and refinement of language models, thereby increasing their ability to understand and help young users with their specific language needs. Our practical application of the French-YMCA corpus focused on AAC techniques, an area where the scarcity of linguistic models that address the communication needs of children with disabilities poses a notable challenge. Through the successful training of adapted next-word prediction models, which outperformed their adult data-based counterparts, we were able to demonstrate our corpus effectiveness in enhancing digital interactions for young individuals. In addition, we are dedicated to continuous research efforts on discovering creative methods to enhance the impact of the French-YMCA corpus. This commitment ensures that our corpus will be accessible online, allowing researchers, developers and educators to easily access it in order to build and evaluate language models suitable for this age group and innovate educational tools.


\section*{Acknowledgements} 
This work was founded by the French national research agency (ANR) in the framework of the AAC4All project (ANR-21-CE19-0051) - \url{https://www.aac4all.org}
\section{Bibliographical References}\label{sec:reference}

\bibliographystyle{lrec_natbib}
\bibliography{lrec-coling2024-example}

\bibliographystylelanguageresource{lrec_natbib}
\bibliographylanguageresource{languageresource}

\end{document}